\pdfoutput=1

\documentclass[11pt]{article}

\usepackage[]{acl}

\usepackage{times}
\usepackage{latexsym}
\usepackage{algorithm}
\usepackage{algorithmic}
\usepackage{graphicx} % Required for inserting images
\usepackage{subcaption}
\usepackage{amsmath}
\usepackage{amssymb}
\usepackage{booktabs}
\usepackage{tabularray}
\usepackage{arydshln}
\usepackage{multicol}

\usepackage[utf8]{inputenc}

\usepackage{microtype}

\title{Asymmetric Bias in Text-to-Image Generation with Adversarial Attacks}

\author{
Haz Sameen Shahgir, Xianghao Kong, Greg Ver Steeg, Yue Dong\\ [3pt]
University of California Riverside\\ [3pt]
\texttt{\{hshah057,xkong016,greg.versteeg,yue.dong\}@ucr.edu}
}

\begin{document}

\maketitle
\thispagestyle{plain}
\pagestyle{plain}

\begin{abstract}
The widespread use of Text-to-Image (T2I) models in content generation requires careful examination of their safety,  including their robustness to adversarial attacks. Despite extensive research on adversarial attacks, the reasons for their effectiveness remain underexplored. This paper presents an empirical study on adversarial attacks against T2I models, focusing on analyzing factors associated with attack success rates (ASR). We introduce a new attack objective - entity swapping using adversarial suffixes and two gradient-based attack algorithms. Human and automatic evaluations reveal the asymmetric nature of ASRs on entity swap: for example, it is easier to replace \textit{``human"} with \textit{``robot"} in the prompt \textit{``a human dancing in the rain."} with an adversarial suffix, but the reverse replacement is significantly harder. We further propose probing metrics to establish indicative signals from the model's beliefs to the adversarial ASR. We identify conditions that result in a success probability of 60\% for adversarial attacks and others where this likelihood drops below 5\%. \footnote{The code and data are available at \url{https://github.com/Patchwork53/AsymmetricAttack}}
\end{abstract}

\section{Introduction}

\begin{figure*}[ht]
\centering
\begin{subfigure}{.332\textwidth}
  \centering
  \includegraphics[width=1\linewidth]{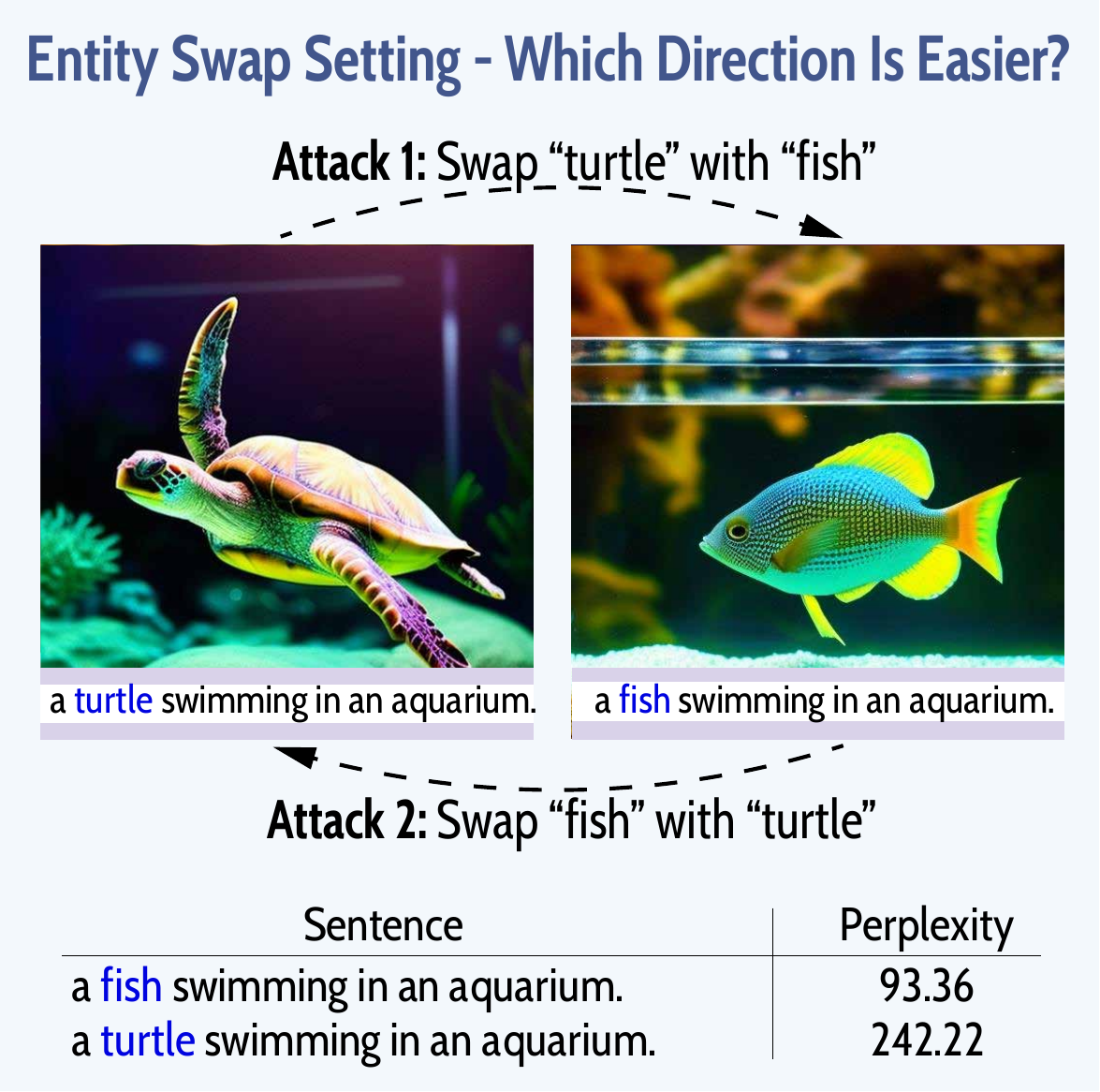}
  \caption{}
  \label{fig:open_fig_a}
\end{subfigure}%
\begin{subfigure}{.37\textwidth}
  \centering
  \includegraphics[width=1\linewidth]{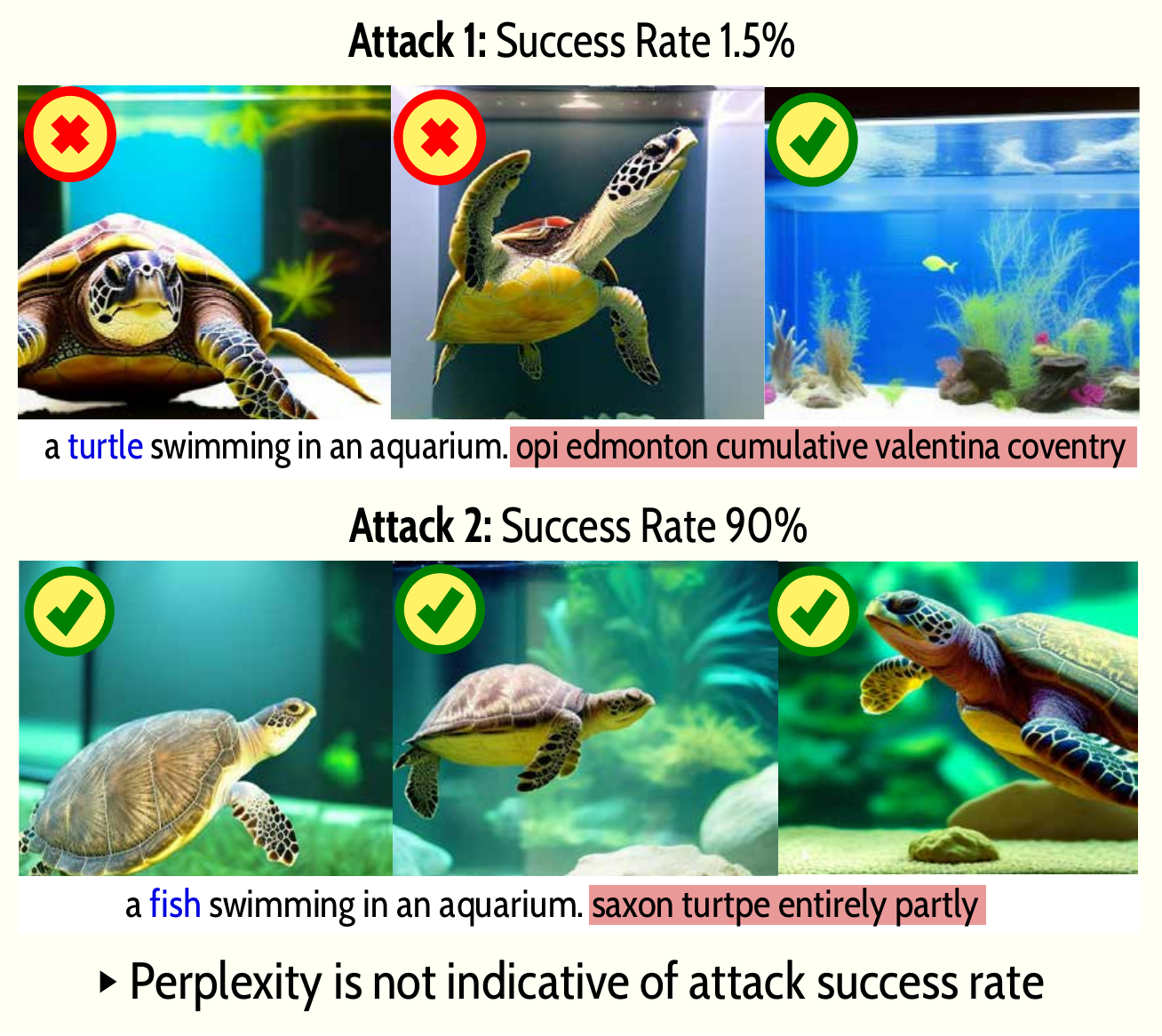}
  \caption{}
  \label{fig:open_fig_b}
\end{subfigure}%
\begin{subfigure}{.285\textwidth}
  \centering
  \includegraphics[width=1\linewidth]{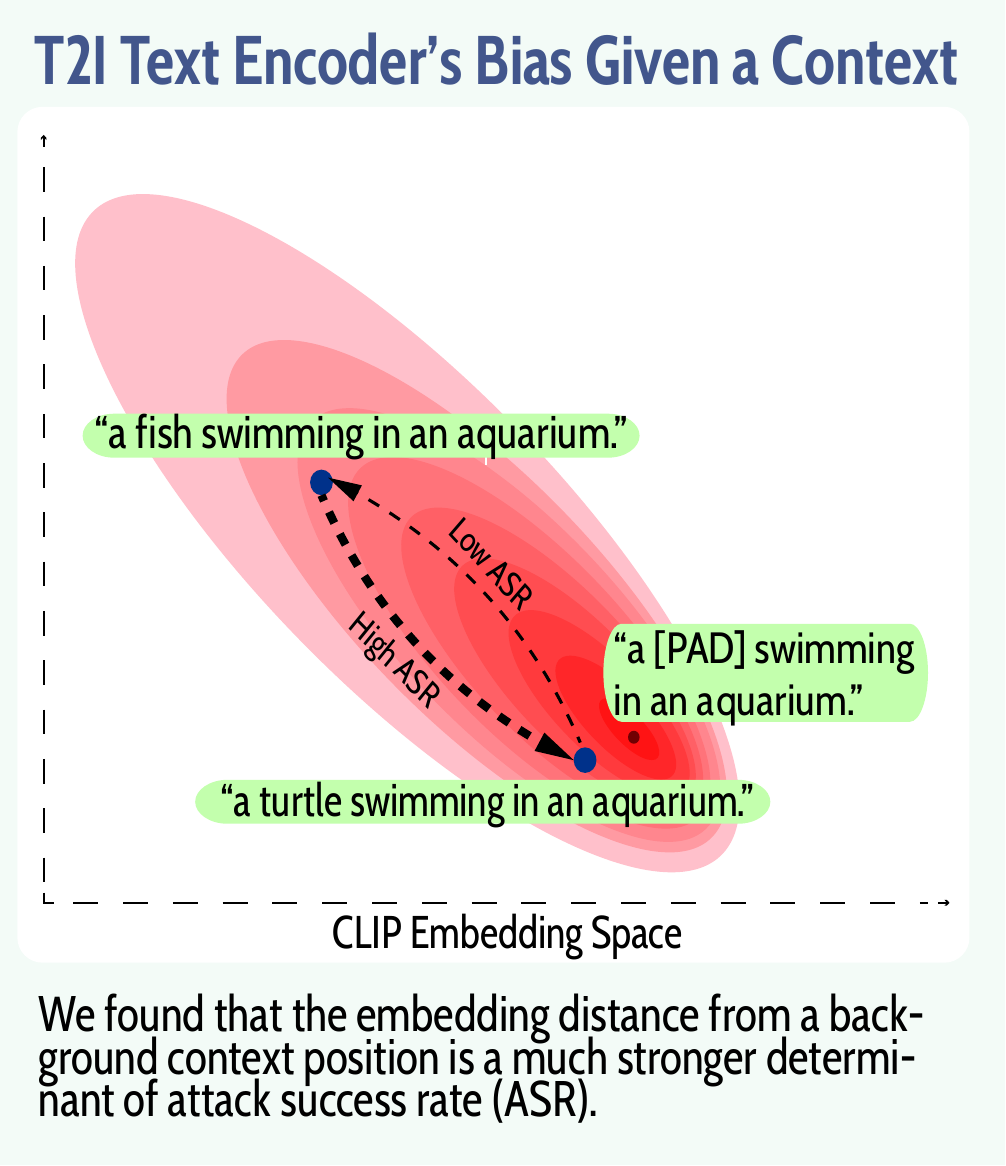}
 \caption{}
 \label{fig:open_fig_c}
\end{subfigure}
\caption{Overview of new attack objective, its asymmetric success rate, and the underlying cause of said asymmetry.}
\label{fig:open_fig}
\end{figure*}

The capabilities of Text-to-Image (T2I) generation models, such as DALL-E 2 \citep{dalle2}, DALL-E 3 \citep{dalle3}, Imagen \citep{imagen} and Stable Diffusion \citep{Rombach_2022_CVPR},   have improved drastically and reached commercial viability.  As with any consumer-facing AI solution, the safety and robustness of these models remain pressing concerns that require scrutiny. 

The majority of research related to T2I safety is associated with the generation of Not-Safe-For-Work (NSFW) images with violence or nudity \citep{qu2023unsafe,rando2022redteam,tsai2023ring}. To counter this, pre-filters that check for NSFW texts and post-filters that check for NSFW images are used \cite{safetychecker}. However, these filters are not infallible \citep{rando2022redteam}, and research into bypassing them, termed `jailbreaking' is advancing \citep{yang2023sneakyprompt,yang2023mma,noever2021reading,fort2023scaling,cliprobustness,maus2023adversarial, qfattack}. These attacks typically view the creation of NSFW-triggering adversarial prompts as a singular challenge, without sufficiently investigating the reasons behind these attacks' effectiveness.

On the other hand, explainability studies have examined the capabilities and shortcomings of text-to-image (T2I) models. They show that T2I models often generate content without understanding the composition ~\citep{kong2023interpretable, west2023generative}, and reveal compositional distractors~\citep{hsieh2023sugarcrepe}.  We identified a specific bias of T2I models linked to adversarial attack success rates, bridging the gap between attack and explainability research. We demonstrate the asymmetric bias of the T2I models by conducting adversarial attacks in a novel entity-swapping scenario, in contrast to the existing setup of removing objects~\cite {qfattack} or inducing NSFW content \cite{yang2023sneakyprompt,yang2023mma}. This setup enables us to investigate the attack success rate in a cyclical setting. 

To study the underlying reasons for the success of adversarial attacks, the attack must be powerful and have a high success rate. This would allow us to ensure that cases with low success rates arise due to the model's internal biases, not simply as a result of the algorithm's shortcomings. We propose two optimizations of existing gradient-based attacks \cite{autoprompt, llmattack} using efficient search algorithms to find adversarial suffix tokens against Stable Diffusion. This approach is based on the observation that existing algorithms for LLM attacks are unnecessarily conservative in generating adversarial perturbations and struggle to efficiently navigate the larger vocabulary size of the T2I text encoder.

Our novel setup and efficient adversarial attack have allowed us to observe an asymmetric attack success rate associated with entity swap. Initially, we hypothesized that long-tail prompts with high perplexity would be more vulnerable to attacks. Surprisingly, we found no strong correlation between the Attack Success Rate (ASR) and the perplexity of the prompt. However, with our proposed measure that evaluates the internal beliefs of CLIP models, we detected indicative signals for ASR, which help identify examples or prompts that are more susceptible to being attacked. Our contributions can be summarized as follows.

\begin{enumerate}
\item We introduce a new attack objective: replacing entities of the prompt using an adversarial suffix. This allows us to study the relation between adversarial attacks and the underlying biases of the model (Figure \ref{fig:open_fig_a}). 

\item We apply an existing gradient-based attack algorithm to execute entity-swap attacks and propose improvements that take advantage of the bag-of-words nature of T2I models. This powerful attack method reveals a clear distinction in the ASR when two entities are swapped in opposite directions, indicating an asymmetry in adversarial attacks (Figure \ref{fig:open_fig_b}).
  
\item We propose a new metric tied to the asymmetric bias of T2I models. This helps us identify vulnerable preconditions and estimate ASR without performing an attack (Figure \ref{fig:open_fig_c}).

\end{enumerate}

\section{Related Works}
\paragraph{Adversarial Attacks} Adversarial attacks, which perturb inputs to cause models to behave unpredictably, have been a long-studied area in the field of adversarial robustness  \citep{szegedy2013intriguing,shafahi2018adversarial,shayegani2023survey}. Previous studies on adversarial attacks focused on discriminative models involving convolutional neural networks \citep{athalye2018synthesizing,hendrycks2018benchmarking}, while recent work has shifted towards examining generative models such as large language models (LLMs) \citep{autoprompt,llmattack,liu2023queryrelevant,mo2023trustworthy,cao2023defending}, Vision Language models (VLMs) \citep{dong2023robust,khare2023understanding,shayegani2023plug}, and Text-to-Image (T2I) models.

\paragraph{Attacks on T2I Models} \citet{qfattack} were among the first to demonstrate that a mere five-character perturbation could significantly alter the generated images. \citet{tsai2023ring} and SneakyPrompt \citep{yang2023sneakyprompt} proposed adversarial attacks using genetic algorithms and reinforcement learning algorithms to perturb safe prompts to generate NSFW content. VLAttack \citep{yin2023vlattack}, MMA-Diffusion \citep{yang2023mma}, and INSTRUCTTA \citep{wang2023instructta} demonstrated that cross-modality attacks can achieve higher success rates than text-only attacks. For defense,  \citet{zhang2023adversarial} proposed Adversarial Prompt Tuning to enhance the adversarial robustness of VLMs. However, to the best of our knowledge, no comparable defense against non-NSFW attacks exists for T2I models.

\paragraph{Vulnerability Analysis} Previous studies \citep{ilyas2019adversarial,shafahi2018adversarial,brown2017adversarial} have explored the reasons for the vulnerability of neural networks to adversarial attacks, especially in image classification. \citet{ilyas2019adversarial} suggested that adversarial examples stem from \textit{non-robust features} in models' representations, which are highly predictive yet imperceptible to humans. \citet{subhash2023universal} suggested that adversarial attacks on LLMs may act like optimized embedding vectors, targeting semantic regions that encode undesirable behaviors during the generation process.

Distinct from previous research, our study analyzes factors in the model's beliefs linked to attack success rates. Unlike prior work focusing on untargeted attacks to trigger NSFW image generations, we introduce a unique entity-swapping attack setup and develop a discrete token-searching algorithm for \textit{targeted attacks}, identifying asymmetric biases in success rates due to the model's internal bias.   Our experiments emphasize the relationship between prompt distributions, model biases, and attack success rates.

\section{Entity Swapping Attack}

This section describes the proposed setup of the entity-swapping attack and the corresponding evaluation metric. Designing a new attack scenario may be straightforward, but developing a suitable measure is not trivial. Towards this end, we propose two efficient discrete token search algorithms for the attack, resulting in improved success rates in entity-swapping attacks.

\subsection{Stable Diffusion}
We study entity-swapping attacks using Stable Diffusion~\citep{Rombach_2022_CVPR}, an open-source \footnote{Licensed under \href{https://huggingface.co/stabilityai/stable-diffusion-xl-base-1.0/blob/main/LICENSE.md}{CreativeML Open RAIL++-M License} for intended for research purposes only.} T2I model based on a denoising diffusion probabilistic model with a U-Net architecture. It uses cross-attention and CLIP~\citep{radfordclip} for text-image alignment and a variational auto-encoder~\citep{kingma2013auto} for latent space encoding. The model's dependence on CLIP text embeddings increases its vulnerability to adversarial attacks. See Appendix \ref{sec:t2i_detail} for more details.

\subsection{Entity Swapping Dataset}
\label{ssec:dataset}

We first constructed datasets with the following key properties to study model bias through entity-swapping attacks.
\begin{enumerate}
\item Each data point should be a pair of sentences - input and target - and T2I models should be able to generate both reliably.
\item The input and target sentences should differ by exactly one noun (i.e., an entity).
\item The input and target sentences must be visually distinct.
\end{enumerate}

As an example, the pair \textit{(``a person in a park.", ``a man in a park.")} satisfies requirements 1 and 2 but not 3.  As our setup for entity-swapping attacks is targeted, namely adversarial attacks need to swap the entities in the images without affecting other parts compared to other attacks that aim to either generate NSFW images or remove objects, we created two datasets to study the effects of adversarial attacks. We manually constructed a small high-quality dataset HQ-Pairs and a larger-scale set derived from an existing dataset MS-COCO.

\paragraph{HQ-Pairs} For the first dataset, we manually crafted 100 pairs for entity-swapping that satisfy all the requirements. We refer to this first dataset as HQ-Pairs (High Quality).

\paragraph{COCO-Pairs}To ensure that our results were not due to selective data selection, we generated a second dataset of 1,000 pairs deterministically from the test split captions of MS-COCO \cite{lin2014microsoft}. We refer to this dataset as COCO-Pairs \footnote{The code to reproduce COCO-Pairs is provided in our codebase.}. Since COCO-Pairs is automatically generated, we attempted to ensure that each data pair satisfies all three requirements. However, generating sentence pairs through stable diffusion and verifying them as visually distinct automatically is not always reliable. We observed some visually non-distinct pairs, such as \textit{(``Herd of zebras ...", ``Images of zebras ...")} within COCO-Pairs despite automatic checks and filtering. See Appendix \ref{sec:gen_coco} for details.

\begin{figure*}[t]
    \centering
    \includegraphics[width=1\linewidth]{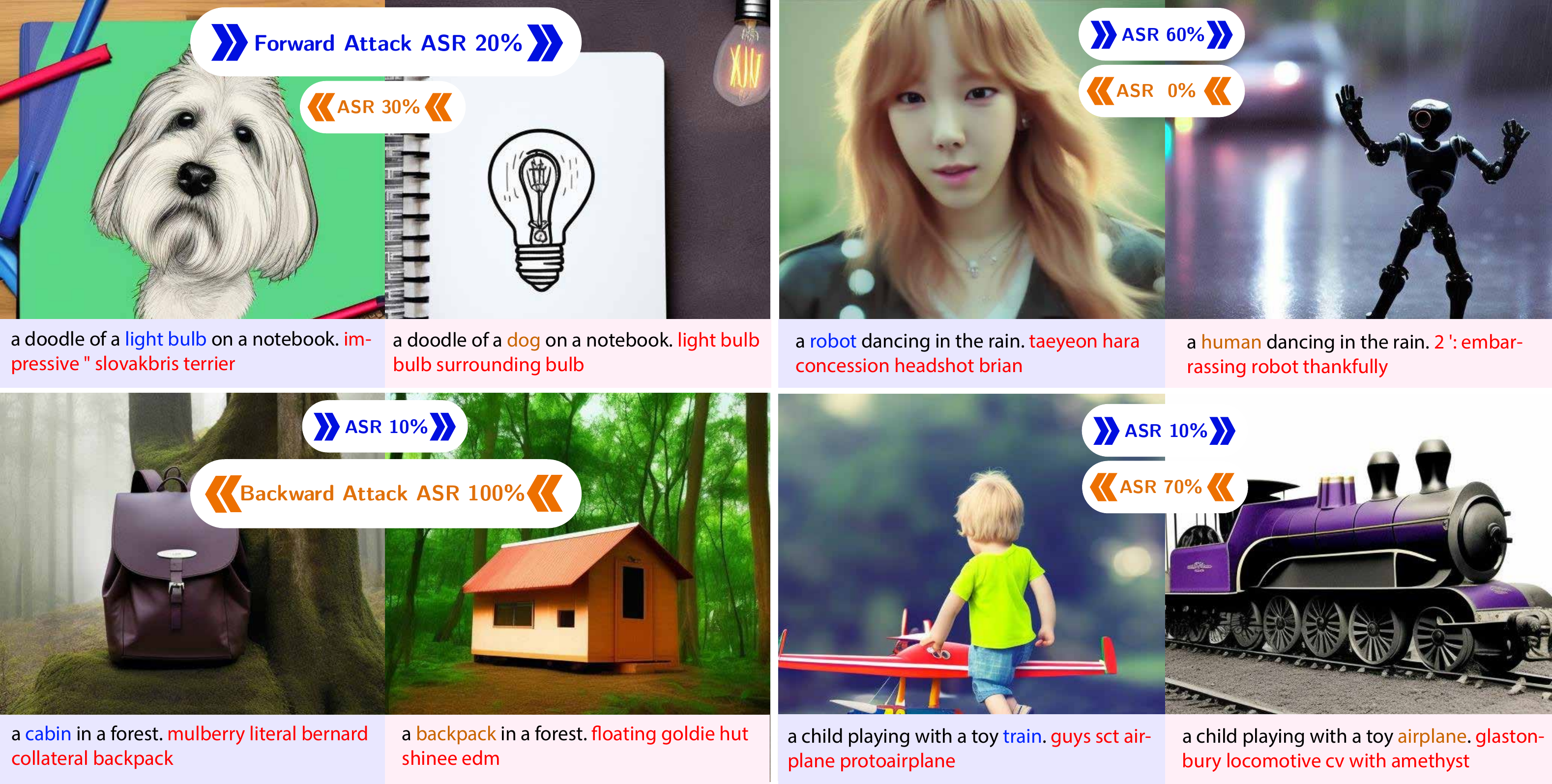}
    \caption{Targeted replacement of entities (blue or orange text) using adversarial suffixes (red highlight) and their corresponding Attack Success rate (ASR) over 10 attack attempts using \href{https://huggingface.co/stabilityai/stable-diffusion-2-1-base}{Stable Diffusion}. This attack setup allows us to study the correlation between prompt distribution and ASR. We observe a clear distinction in ASR when performing entity-swapping with reversed directions. The rest of the paper explores explanations and measures that can detect and predict ASR without performing the attack itself.
    } 
    \label{fig:attack_examples}
\end{figure*}

\subsection{Proposed Attack}

We examine how the underlying data distribution of prompts influences the success rate of entity-swapping attacks on T2I models. Our approach is straightforward: rather than manipulating T2I to produce NSFW images or completely removing an object, we aim to replace an object in the image with another targeted one. This approach also allows us to explore the feasibility of reverse attacks by inserting adversarial tokens. Examples of our attack setup can be found in Figure \ref{fig:attack_examples}.

The CLIP text-encoder transforms prompt tokens $x_{1:n}$ into $n$ hidden states with dimension $D$. Let the operation $\mathcal{H}$ represent the combined process of encoding tokens $x_{1:n}$ and reshaping the hidden states into a vector of length $n\times D$. 

\begin{equation}
    \mathcal{H}(x_{1:n})  = \text{Flatten}(\text{CLIP}(x_{1:n}))
\end{equation}

Our attack targets the CLIP embedding space and aims to maximize a score function that measures the shift from the input token embeddings $\mathcal{H}(x^T_{1:n})$ towards the target token embeddings $\mathcal{H}(x^S_{1:n})$ using cosine similarity:

\begin{equation}
\label{eqn:score_eqn}
\begin{split}
   \mathcal{S}(x_{1:n})=w_t\times\text{cos}(\mathcal{H}(x^T_{1:n}),\mathcal{H}(x_{1:n}))-\\w_s\times\text{cos}(\mathcal{H}(x^S_{1:n}),\mathcal{H}(x_{1:n})) 
\end{split}
\end{equation}

Optimizing $\mathcal{S}$ is challenging due to the discrete token set and the exponential search space ($k^{|V|}$ for $k$ suffix tokens), making simple greedy search intractable. Current solutions based on HotFlip \cite{hotflip} and concurrent work applied to Stable Diffusion \cite{yang2023mma}, take gradients w.r.t. one-hot token vectors and replace tokens for all positions in the suffix simultaneously.  The linearized approximation of replacing the $i^{th}$ token, $x_i$, is computed by evaluating the following gradients:

\begin{equation}
\label{eqn:L_diff_eqn_main}
\nabla_{e_{x_i}}\mathcal{L}(x_{1:n}) \in \mathbb{R}^{|V|}, \quad \mathcal{L}(x_{1:n})=-\mathcal{S}(x_{1:n})
\end{equation}
where ${e_{x_i}}$ denotes the one-hot vector representing the current value of the $i^{th}$ token.

\subsection{Proposed Optimization Algorithms}
Based on existing gradient-based methods \cite{llmattack,autoprompt}, we propose two efficient algorithms to find adversarial suffix tokens against Stable Diffusion.

\paragraph{Single Token Perturbation}
This is a straightforward modification of the Greedy Coordinate Gradient algorithm \citep{llmattack} using our loss function defined in Eqn. \ref{eqn:L_diff_eqn_main}. At each optimization step, our algorithm selects $k$ tokens with the highest negative loss as replacement candidates, $\chi_i$, for each adversarial suffix position $i$. It then creates $B$ new prompts by randomly replacing one token from the candidates. Each prompt in $B$ differs from the initial prompt by only one token. The element of B with the highest $\mathcal{S}$ is then assigned to $x_{1:n}$. We repeat this process $T$ times.

\paragraph{Multiple Token Perturbation}
 Unlike the LLMs targeted by \citet{llmattack}, CLIP models operate more like bag-of-words \citep{bagofwordsvlm} without capturing semantic and syntactical relations between words. Furthermore, Genetic Algorithms \citep{sivanandam2008genetic} have proved effective on Stable Diffusion \citep{qfattack, yang2023sneakyprompt} for generating adversarial attacks. Inspired by this apparent weakness in CLIP models, we hypothesized that replacing multiple tokens simultaneously could improve the convergence speed.

In detail, the algorithm selects $k$ tokens and creates $B$ new prompts by randomly replacing multiple token positions. Drawing inspiration from the classic \textit{exploration versus exploitation} strategy in reinforcement learning \citep{sutton2018reinforcement}, we initially replace all tokens and then gradually decrease the replacement rate to $25\%$. Figure \ref{fig:attack_examples} illustrates some adversarial suffixes generated using this algorithm. Details of both algorithms are provided in the Appendix \ref{sec:algorithm}.

\begin{figure}[ht]
    \centering
    \includegraphics[width=1\linewidth]{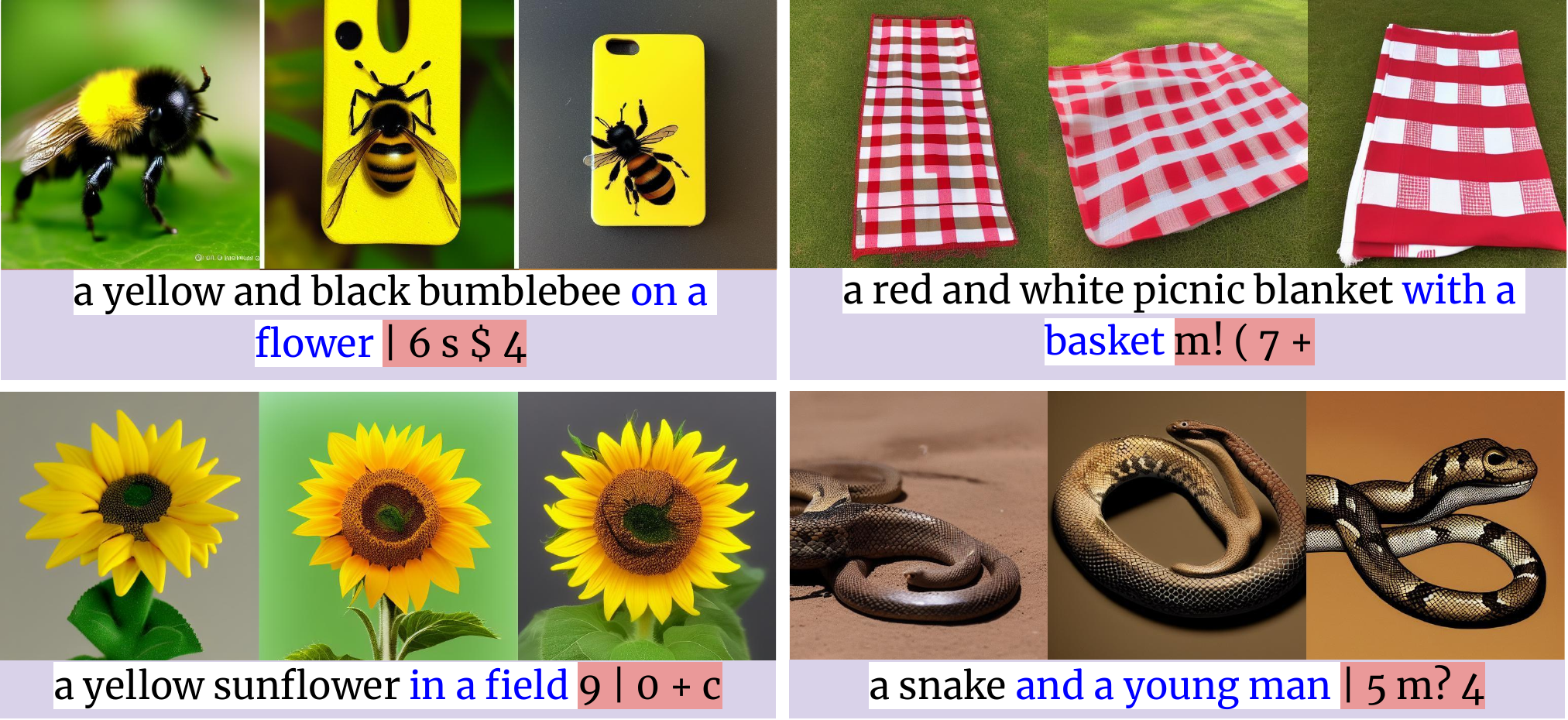}
    \caption{The emulation of restricted token attack (untargeted) from \citet{qfattack} using five ASCII tokens with \href{https://huggingface.co/CompVis/stable-diffusion-v1-4}{Stable Diffusion 1.4}. The blue text indicates the part we want to remove. We set $w_t=0$ in Eqn. \ref{eqn:score_eqn}.}
    \label{fig:qf_emulate}
\end{figure}

\paragraph{Token Restrictions}
For finer control over token search, we can limit the adversarial suffix to a set of tokens $\mathcal{A}$. By setting the gradients of the $V - \mathcal{A}$ tokens to infinity before the Top-k operation, we ensure only $\mathcal{A}$ tokens are chosen. This method allows us to mimic QFAttack \cite{qfattack}, as shown in Figure \ref{fig:qf_emulate}, or generate undetectable attacks by excluding target synonyms in the attack suffix.

\subsection{Proposed Attack Evaluation}
\label{subsec:eval_def}
To assess the success of a targeted entity-swapping attack, we use a classifier to verify if the generated image matches the input or target prompt. Given a tuple \textit{(input text, target text, generated image)}, we define a classifier $\mathcal{C}$ as follows:

\begin{equation}
\begin{split}
\mathcal{C}(\textit{input text}, \textit{target text}, \textit{generated image})\\ =
\begin{cases} 
+1 & \text{if image matches target text} \\
-1 & \text{if image matches input text} \\
0 & \text{otherwise}.
\end{cases}
\end{split}
\label{per_img_success}
\end{equation}

When trying to change \textit{``A backpack in a forest''} to \textit{``A cabin in a forest''}, we noticed that some of the generated images depicted \textit{``People in a forest''} or  \textit{``A cabin and a backpack in a forest''} instead. We define such cases as class $0$. Class $+1$ alone indicates a successful attack, but this three-class framework enables a more comprehensive comparison between human judgments and our proposed classifiers.

\paragraph{Attack Success Rate (ASR)} We define an adversarial suffix as \textit{ successful} if the target text is a suitable caption for the majority of images generated by an attack prompt using a T2I model. For example, if we generate 5 images with an appended adversarial suffix prompt \textit{``A backpack in a forest.\colorbox{pink}{titanic tycoon cottages caleb dojo}''}, we will consider the adversarial suffix \textbf{successful} if 3 or more images match the target prompt  \textit{``A cabin in a forest''}. 

\paragraph{Human Evaluations/Labels} We gather evaluations from three human evaluators \footnote{Our evaluations were conducted by three non-author, native English-speaking volunteers who generously offered their time without compensation. We sincerely thank them for their commitment and good faith effort in labeling. } for 200 random samples by presenting them a WebUI (Appendix \ref{sec:webui}) with the generated image and two checkboxes for input text and target text. They are instructed to select texts that match the image and can select one, both, or neither, i.e. into three classes as established in Eqn. \ref{per_img_success}. The $\text{Gwet-AC}_1$ metric \cite{gwet2014handbook} of the three evaluators is $0.765$ and the pairwise Cohen's Kappa $\kappa$  metrics \cite{cohen1960coefficient} are $0.659, 0.736$, and $0.779$, indicating a high degree of agreement. We consider the majority vote among evaluators as ground truth.

\paragraph{Choice of the Classifier} We generate multiple attack suffixes for each input-target pair to determine attack success rates. Due to the large volume of images, we employ human evaluators for a subset and VLM-based classifiers for the full set evaluation. We test InstructBLIP \cite{instructblip}, LLaVA-1.5 \cite{llava15} and CLIP \cite{radfordclip}, and compare their performance with human labels.

\begin{table}[ht]
\centering
\resizebox{0.4\textwidth}{!}{
\centering
\begin{tabular}{l|ccc}
\toprule
   Model &  \# Classes &  Accuracy &   F1 \\
\midrule
\textbf{InstructBLIP} &     \textbf{3} &      \textbf{0.79} & \textbf{0.75} \\
LLaVA-1.5 &          3 &      0.76 & 0.74 \\
CLIP &          3 &      0.62 & 0.55 \\
CLIP-336 &          3 &      0.60 & 0.55 \\
\midrule
\textbf{InstructBLIP} &       \textbf{2} &      \textbf{0.86} & \textbf{0.84} \\
LLaVA-1.5 &          2 &      0.83 & 0.81 \\
CLIP &          2 &      0.70 & 0.69 \\
CLIP-336 &          2 &      0.68 & 0.67 \\
\bottomrule
\end{tabular}
}
\caption{Comparison of Automated Evaluation Models. \# Classes = $3$ means the model outputs are categorized into classes $\{-1,0 \text{ and } 1\}$ as defined in Eqn. \ref{per_img_success}. Since classes $\{-1, 0\}$ both correspond to unsuccessful attacks, we collapse them into a single class $0$ and report the performance of the VLM models with \# Classes = $2$.}
\label{tab:vlm_performance}
\end{table}

\begin{figure*}[t!]
    \centering
    \includegraphics[width=0.97\linewidth]{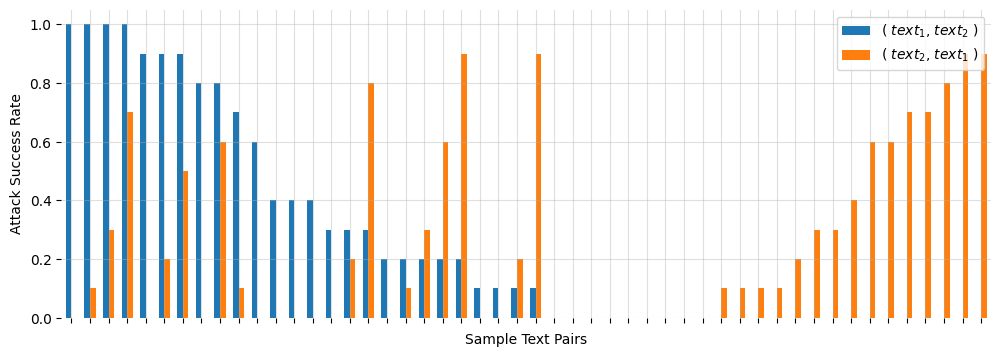}
    \caption{Comparison of pair-wise attack success rate on HQ-Pairs using  Multiple Token Perturbation Algorithm.}
    \label{fig:assymetric_success_rate}
\end{figure*} 

For InstructBLIP and LLaVA-1.5, we use the prompt \textit{`Does the image match the caption [PROMPT]? Yes or No?'}. For CLIP models, an image is classified as $+1$ if its target text similarity is above $1-\gamma$ and its input text similarity is below $\gamma$ and $-1$ for the reverse case. All other cases are classified as $0$. Table \ref{tab:vlm_performance} shows the agreement of different automatic classifiers with ground truths from our human evaluators. We use the optimal threshold $\gamma$ ($\gamma_{CLIP}=0.0034$ and $\gamma_{CLIP-336}=0.0341$) that maximizes the F1 score. Since InstructBLIP shows the best alignment with human evaluation, we use InstructBLIP as our sole classifier in subsequent sections.

\section{Experiments and Results}
\label{sec: results}
This section presents the experimental details and results of adversarial attacks for entity-swapping, involving the insertion of adversarial suffixes.

\subsection{Experimental Setups}

We evaluate \href{https://huggingface.co/stabilityai/stable-diffusion-2-1-base}{Stable Diffusion v2-1-base} on the HQ-Pairs dataset of 100 input-target pairs to compare the effectiveness of Single and Multiple Token Perturbation. We run each algorithm 10 times per pair with $T=100$ steps with $k=5$ and $B=512$, which yields 10 adversarial attacks per pair, and we generate 5 images per attack. The two algorithms are evaluated against each other on $100\times 10 \times 5=5000$ generated images. We set $w_t=w_s=1$ in Eqn. \ref{eqn:score_eqn} for the experiments. Afterward, we evaluate COCO-Pairs (1000 pairs) using the Multiple Token Perturbation algorithm to establish the asymmetric bias phenomenon with the same hyperparameters. We used a single Nvidia RTX 4090 GPU for all experiments, including attack, image generation, and automated evaluation, totaling around 500 GPU hours.

\subsection{Overall Attack Results}
Using the same hyperparameters and compute budget, our Multiple Token Perturbation algorithm outperforms the Single Token Perturbation ( ASR $26.4\%$ vs. $24.4\%$ for 1000 attacks). \citet{llmattack} showed that Single Token Perturbation was an effective adversarial suffix-finding strategy for LLMs. However, the CLIP text is relatively lightweight compared to LLMs and behaves more like a bag-of-words model \cite{bagofwordsvlm}. CLIP also has a larger vocabulary compared to LLMs ($~50K$ vs. $32K$) which leads to a larger unrestricted search space ($\sim10^{24}$ vs. $\sim10^{23}$ for 5 token suffixes). We find that updating multiple tokens at each time step leads to faster convergence, likely because CLIP demonstrates a reduced emphasis on the semantic and syntactical relationships between tokens. Our findings corroborate the effectiveness of the Genetic Algorithm in \citet{qfattack}, which resembles multiple token perturbations but operates in an untargeted setting without a gradient-based algorithm. \textit{We employ Multiple Token Perturbation for all subsequent experiments}.

\subsection{Forward and Backward Attack Results}
\label{subsec:assymetric_results}

One of our key findings is the strong asymmetry of adversarial attack success rate, as illustrated in Figure \ref{fig:assymetric_success_rate}. For instance, attacks from \textit{`A swan swimming in a lake.'} to \textit{`A horse swimming in a lake.'} failed in all ten attempts, whereas the reverse direction achieved an ASR of 0.9. In other cases, the forward and backward ASRs aren't inversely proportional. For example, both directions between \textit{`A man reading a book in a library.'} and \textit{`A woman reading a book in a library.'} have moderate ASRs of 0.7 and 0.5, respectively, while pairs like (\textit{`A dragon and a treasure chest.'}, \textit{`A knight and a treasure chest.'}) fail in both directions. Inspired by these asymmetric observations, we conducted further experiments to analyze the relationship between prompt distribution and attack success rate.

\section{Asymmetric ASR Analysis}
\label{sec:asr_analysis}
This section discusses our experiments to analyze the asymmetric ASR observed in Section \ref{subsec:assymetric_results}. We aim to investigate the model's internal beliefs that may lead to these distinct attack success rate (ASR) differences from opposite directions.  We propose three potential factors for this asymmetry: the difficulty of generating the target text (BSR, Eqn. \ref{eqn:base_succ}), the \textit{naturalness} of the target text relative to the input text ($\Delta_1$, Eqn. \ref{eqn:ppl_diff}), and the difference in distance from the target text to the baseline compared to that from the input text ($\Delta_2$, Eqn. \ref{eqn:baseline_diff}).

\subsection{Probe Metrics}

We initially speculated that ASR might be related to the difficulty in generating the target prompt, leading us to evaluate the Base Success Rate (BSR) of target generation. 
\begin{equation}
    \text{BSR}=\frac{\text{Successful Generations}}{\text{Generation Attempts}}
\label{eqn:base_succ}
\end{equation}

BSR assesses the T2I model's ability to generate an image that matches the input prompt without any adversarial suffixes. Stable Diffusion is often unable to generate novel compositions not present in its training data \cite{west2023generative} and struggles with generating co-hyponym entities in the same scene \cite{tang2022daam}. We find that even simple scenes such as \textit{ ``A dragon guarding a treasure.''} are inconsistently produced (See Appendix \ref{sec:four_plots} for examples). Therefore, if the T2I models struggle with the target alone, adversarial attacks aimed at generating them are likely to be even more challenging.

\begin{figure}[ht]
    \centering
    \includegraphics[width=1\linewidth]{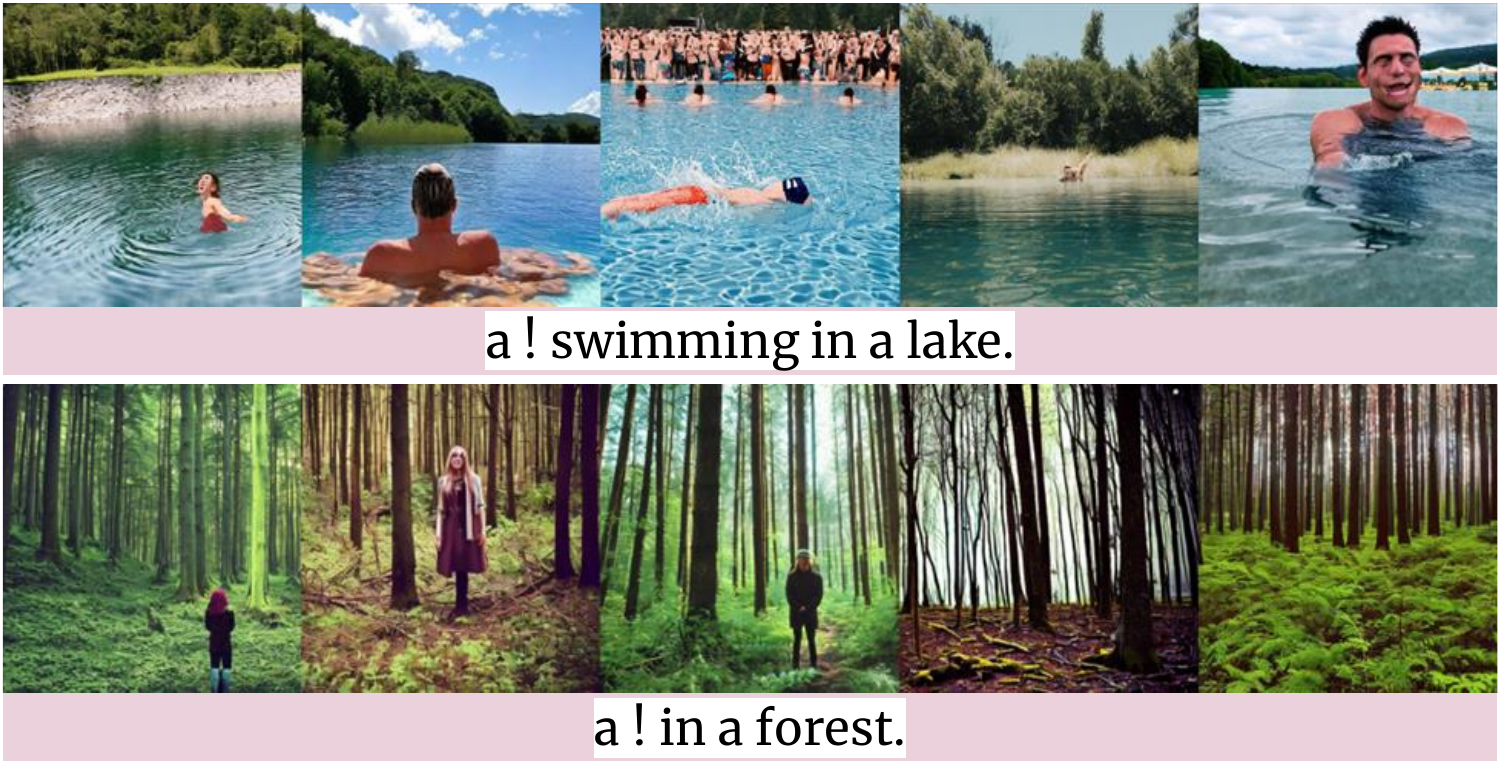}
    \caption{Baseline Distance Difference measures the inherent biases of T2I models. This can be observed by prompting Stable Diffusion a PAD token in place of an entity.}
    \label{fig:xample}
\end{figure}

\begin{figure*}[t!]
\centering
\begin{subfigure}{.6\textwidth}
  \centering
  \includegraphics[width=1\linewidth]{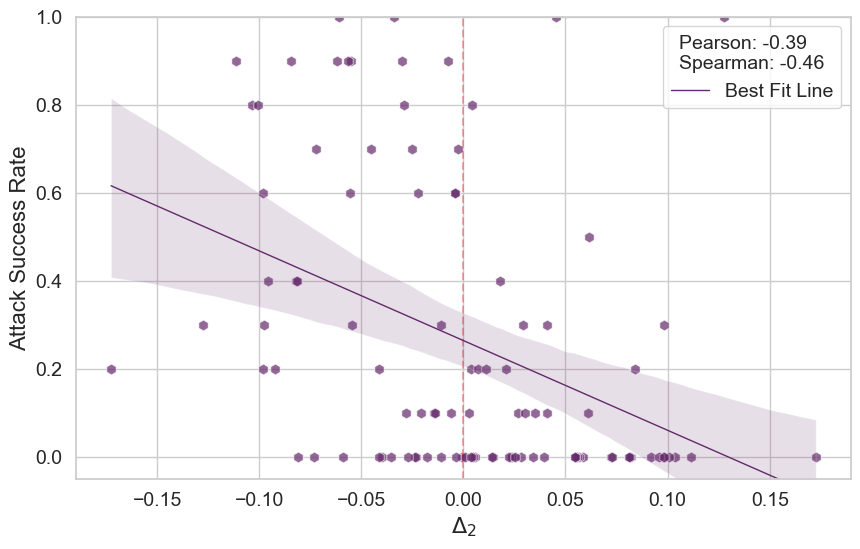}
   \caption{ASR vs. Baseline Distance Difference ($\Delta_2$ in Eqn. \ref{eqn:baseline_diff})}
  \label{fig:f2}
\end{subfigure}%
\begin{subfigure}{.4\textwidth}
  \centering
  \includegraphics[width=0.95\linewidth]{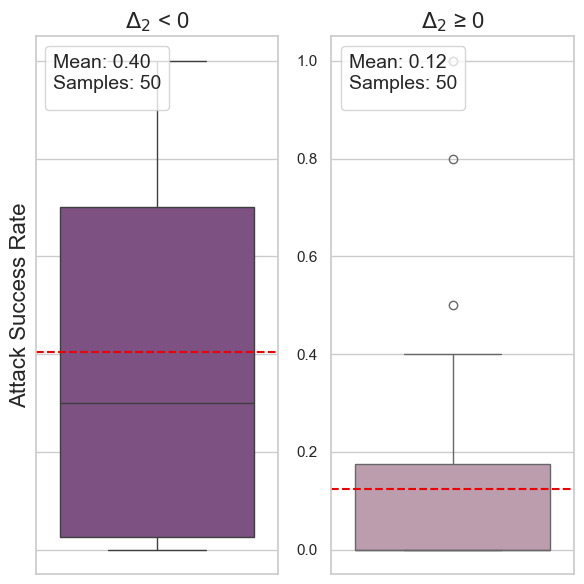}
  \caption{ASR for Negative and Positive $\Delta_2$ }
  \label{fig:f4}
\end{subfigure}
\caption{Correlation of ASR with Baseline Distance Difference $\Delta_2$. Data is reported using the Multiple Token Perturbation algorithm on HQ-Pairs. $\Delta_2$ shows a moderate negative correlation with ASR.}
\label{fig:assymetric_ASR}
\end{figure*}

We also speculated that the difference in Perplexity $\Delta_1$, measuring \textit{how natural or plausible a prompt is}, might be associated with asymmetric ASR. For example, \textit{ ``A swan swimming in a lake''} is a more natural scene than \textit{``A horse swimming in a lake''}. Using \texttt{text-davinci-003} by OpenAI \cite{gpt3}, we calculate the perplexity difference

\begin{equation}
    \mathbf{\Delta_{1}}(x^T_{1:n},x^S_{1:n}) = \text{PPL}(x^T_{1:n}) - \text{PPL}(x^S_{1:n}).
\label{eqn:ppl_diff}
\end{equation}
where $\text{PPL}(x_{1:n}) = e^{-\frac{1}{n}\sum_{i=1}^{n} \log P(x_i | x_{1:i-1})}$ is the perplexity for the sequence $x_{1:n}$.
\\\\
Furthermore, we introduce a new metric termed \textbf{\textit{Baseline Distance Difference}}, denoted as \textbf{$\Delta_2$}.  Figure \ref{fig:xample} shows that T2I models have inherent biases towards certain objects. We denote this phenomenon as the \textit{baseline} - answering \textit{what would Stable Diffusion generate if prompted with ``A \_\_\_\_\_ swimming in a lake"}. Intuitively, targets closer to this baseline should be easier to generate.

\begin{equation}
\begin{split}
      \mathbf{\Delta_{2}}(x^T_{1:n},x^S_{1:n}) = \text{cos}(\mathcal{H}(x^T_{1:n}),\mathcal{H}(x^B_{1:n})) \\- \text{cos}(\mathcal{H}(x^S_{1:n}),\mathcal{H}(x^B_{1:n})).
\end{split}
\label{eqn:baseline_diff}
\end{equation}

\subsection{Results}

We generated 64 images for each sentence in HQ-Pairs and COCO-Pairs. We counted the number of successful generations to determine the BSR as defined in Eqn. \ref{eqn:base_succ}.

On the HQ-Pairs dataset, we find that Perplexity Difference $\Delta_1$ has a negligible correlation with ASR (Pearson $r=0.05$ and Spearman $\rho=-0.06$). This is counterintuitive because we expected that a target with lower perplexity compared to the input text would be easier to generate through an adversarial attack. We also observed that ASR has a weak positive correlation with BSR (Pearson $r=0.28$ and Spearman $\rho=0.38$) and a moderate correlation with $\Delta_2$ (Pearson $r=-0.39$ and Spearman $\rho=-0.46$. See Figure \ref{fig:f2}). In particular, Figure \ref{fig:f4} shows that the mean ASR is 0.40 when $\Delta_2$ is negative, while it drops to just 0.12 when $\Delta_2$ is positive. Thus, $\Delta_2$ allows us to estimate, to some extent, the probability of a successful adversarial attack. We present more correlation plots of ASR with Perplexity Difference and BSR in Appendix \ref{sec:four_plots}.

\subsection{Predictor for Successful Attack}
\label{subsec:predictors}
Considering the observed correlations of BSR (of the target text) and $\Delta_2$ with attack success rates, this section explores whether the combination of these two indicators can predict the probability of a successful entity-swapping attack.

\begin{table}[ht]
\centering
\resizebox{0.48\textwidth}{!}{

\begin{tabular}{c c |c c| c c}
\toprule
 & & \multicolumn{2}{c}{HQ-Pairs}  & \multicolumn{2}{|c}{COCO-Pairs}  \\

BSR & $\Delta_2$ & Num. & Avg. ASR & Num. & Avg. ASR \\
\midrule
Low     &  Neg.     & 23  & 0.174  & 260  & 0.129 \\
Low    &  Pos.  & 19  & 0.047  & 274& 0.087 \\
\textbf{High}  & \textbf{Neg.}      & \textbf{27}  & \textbf{0.6} & \textbf{239}& \textbf{0.349}\\
High  & Pos.   & 31  & 0.171& 226& 0.213\\
\midrule
All & All & 100 & 0.264 & 1000 & 0.189\\
\bottomrule
\end{tabular}
}
\caption{Average ASR for different combinations of BSR and $\Delta_2$ on COCO-Pairs dataset. We define $\text{BSR}\geq0.9$ as \textit{high}. The average BSR of the target text of HQ-Pairs and COCO-Pairs were 0.82 and 0.698 respectively.}
\label{tab:condition_categories}
\end{table}

Table \ref{tab:condition_categories} shows that our probe metric acts as a reliable predictor of attack success: when BSR (of the target text) is high and $\Delta_2$ is negative for a given input-target text pair, adversarial attacks have a 60\% chance of success on the HQ-Pairs dataset, compared to only 5\% when BSR is low and $\Delta_2$ is positive. Thus, considering both BSR and $\Delta_2$ together enhances the prediction accuracy of an attack's success likelihood. We further validate our findings on the much larger COCO-Pairs dataset. Although the differences are not as pronounced as those in the HQ-Pairs, due to limitations explained in Section \ref{ssec:dataset}, we still observe that high BSR and negative $\Delta_2$ remain indicative of a higher likelihood of successful adversarial attacks.
We also identified factors akin to existing research on general elements associated with attack success rates, like the length of the adversarial suffix. These factors, together with our experimental results, are detailed in Appendix \ref{sec:other_factors}.

\section{Conclusion}
This paper presents an empirical study on adversarial attacks targeting text-to-image (T2I) generation models, with a specific focus on Stable Diffusion. We define a new attack objective: entity-swapping, and introduce two gradient-based algorithms to implement the attack. Our research has identified key factors for successful attacks, revealing the asymmetric nature of attack success rates for forward and backward attacks in entity-swapping. Furthermore, we propose probing metrics to associate the asymmetric attack success rate with the asymmetric bias within the T2I model's internal beliefs, thus establishing a link between a model's bias and its robustness against adversarial attacks.

\section{Limitations}

Our analysis establishes the asymmetric bias phenomenon for Stable Diffusion but whether all T2I models have such bias is an open question. Closed-source T2I models with different architectures such as Imagen and DALL$\cdot$E may be immune to the asymmetric bias phenomenon or their creators may have mitigated biases through careful data curation.

One of our key findings is that asymmetric bias is not intuitive. Although humans might consider ``fish" to be a more natural option (and likely more abundant in the training data) for ``A \_\_\_\_ in an aquarium", we find that Stable Diffusion is strongly biased towards ``turtle" instead. We leave exploring the underlying reason for this non-intuitive bias as future work.

We observed that gradient-based algorithms tend to include the target word in the adversarial suffix. Concurrent works that aim to generate undetectable NSFW attacks use a dictionary to prevent this. Since we target benign words and have different targets for every attack, we could not use a similar approach. We explore explicitly forbidding tokens corresponding to the target word, but the algorithm still finds synonyms or different tokenizations of the target word. Forbidding the target word proved to be a nontrivial and ultimately, we did not consider generating true adversarial attacks to be a central focus of our investigation of model bias. Another technical challenge is the need to compute BSR which involves generating a statistically significant number of images (64 in our experiments) for the same prompt. Finding ways to approximate the BSR is an area for future research.

% Entries for the entire Anthology, followed by custom entries
\bibliography{anthology,custom}
\bibliographystyle{acl_natbib}

\onecolumn
\appendix
\label{sec:appendix}

\section{Generating COCO-Pairs}
\label{sec:gen_coco}
Starting from 5000 captions, we filter out long captions and use a Named-Entity-Recognition model \cite{Nadeau2007ASO} to identify the first noun in the sentence and use a Fill-Mask model \cite{devlin2018bert} to replace it with another noun. We use the NLTK \cite{loper2002nltk} library and several heuristics to prevent synonyms, hyponym-hypernym, and nonvisualizable nouns from being selected. We are left with 2093 \textit{(base caption, synthetic caption)} pairs, from which we sample 500. This yields 1000 sentence pairs in total by considering both directions.

\section{Algorithms}
\label{sec:algorithm}

\begin{algorithm}
\caption{Single Token Perturbation}
\begin{algorithmic}[1]
\REQUIRE Initial prompt $x_{1:n}$, modifiable subset $I$, iterations $T$, loss $\mathcal{L}$, score $\mathcal{S}$, batch size $B$, $k$
\FOR{$t \in T$}
    \FOR{$i \in I$}
        \STATE $\chi_i \leftarrow \text{Top-}k(-\nabla_{x_i} \mathcal{L}(x_{1:n}))$
        \COMMENT{Compute top-$k$ promising token substitutions}
    \ENDFOR
    \FOR{$b = 1, \ldots, B$}
        \STATE $x_{1:n}^{(b)} \leftarrow x_{1:n}$
        \COMMENT{Initialize element of batch}
        \STATE $x_i^{(b)} \leftarrow \text{Uniform}(\chi_i)$, where $i \leftarrow \text{Uniform}(I)$
        \COMMENT{Select random replacement token}
    \ENDFOR
    \STATE $x_{1:n} \leftarrow x_{1:n}^{(b^*)}$, where $b^* = \arg\max_b \mathcal{S}(x_{1:n}^{(b)})$
    \COMMENT{Compute best replacement}
\ENDFOR
\ENSURE Optimized prompt $x_{1:n}$
\end{algorithmic}
\label{alg:single_token}
\end{algorithm}

\begin{algorithm}
\caption{Multiple Token Perturbation}
\begin{algorithmic}[1]
\REQUIRE \textbf{Input:} Initial prompt $x_{1:n}$, modifiable subset $I$, iterations $T$, loss $\mathcal{L}$, score $\mathcal{S}$, batch size $B$, $k$, $\epsilon_f$, $\epsilon_s$
\STATE $\epsilon \leftarrow \epsilon_s$
\FOR{$t \in T$}
    \FOR{$i \in I$}
        \STATE $\chi_i \leftarrow \text{Top-}k(-\nabla_{x_i} \mathcal{L}(x_{1:n}))$
        \COMMENT{Compute top-$k$ promising token substitutions}
    \ENDFOR
    \FOR{$b = 1, \ldots, B$}
        \STATE $x_{1:n}^{(b)} \leftarrow x_{1:n}$
        \COMMENT{Initialize element of batch}
        \FOR{$i \in I$}
            \IF{$\mathcal{P}(\epsilon)$}
                \STATE $x_i^{(b)} \leftarrow \text{Uniform}(\chi_i)$
                \COMMENT{Select random replacement token}
            \ENDIF
        \ENDFOR
    \ENDFOR
    \STATE $x_{1:n} \leftarrow x_{1:n}^{(b^*)}$, where $b^* = \arg\max_b \mathcal{S}(x_{1:n}^{(b)})$
    \COMMENT{Compute best replacement}
    \STATE $\epsilon \leftarrow \max(\epsilon_f, \epsilon_s - \frac{t}{T})$
    \COMMENT{Reduce the replacement probability}
\ENDFOR
\ENSURE \textbf{Output:} Optimized prompt $x_{1:n}$
\end{algorithmic}
\label{alg:multi_token}
\end{algorithm}

\clearpage
\section{Additional Examples of Asymmetric Bias}
\begin{table}[!h]

\begin{tblr}{
  colspec = {X[-1,l]X[-1,c]X[-1,c]X[-1,c]X[-1,c,h]},
}
\hline
\textbf{Sentence Pair (\textcolor{blue}{1} / \textcolor{orange}{2})} & $\boldsymbol{\Delta_2}$ & \textbf{ASR} $\mathbf{\textcolor{blue}{1}\rightarrow\textcolor{orange}{2}}$ & \textbf{ASR} $\mathbf{\textcolor{orange}{2}\rightarrow\textcolor{blue}{1}}$ & \textbf{Example} \\
\hline
a (\textcolor{blue}{plane} / \textcolor{orange}{hot air balloon}) in the sky at sunset. & -0.1 & \textbf{80\%} & 0\%	& \includegraphics[width=0.17\textwidth]{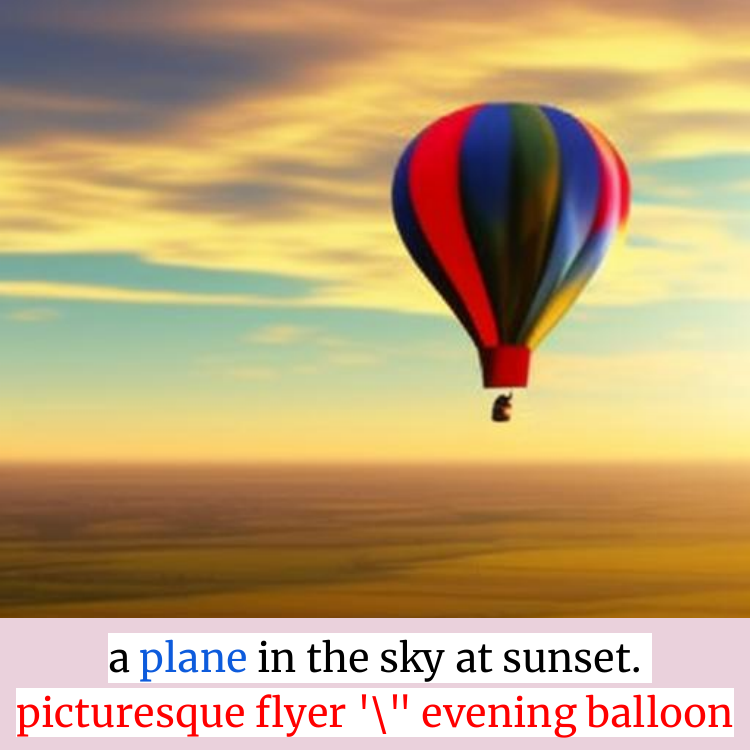}\\

a (\textcolor{blue}{cabin} / \textcolor{orange}{backpack}) on a mountain. & -0.08 & \textbf{90\%} & 20\%		&\includegraphics[width=0.17\textwidth]{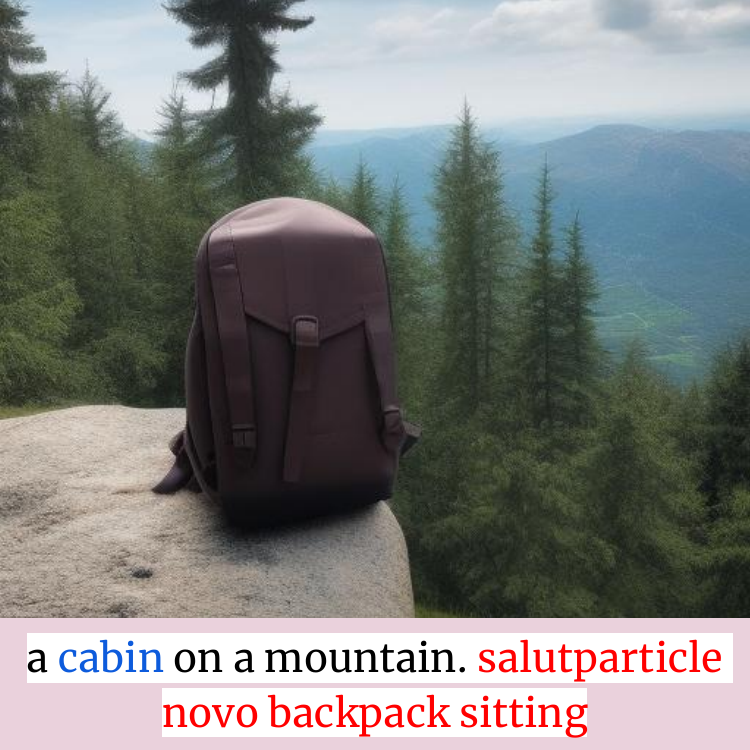}\\

an owl in a (\textcolor{blue}{forest} / \textcolor{orange}{shopping mall}). & -0.07 & \textbf{70\%} & 0\%	&\includegraphics[width=0.17\textwidth]{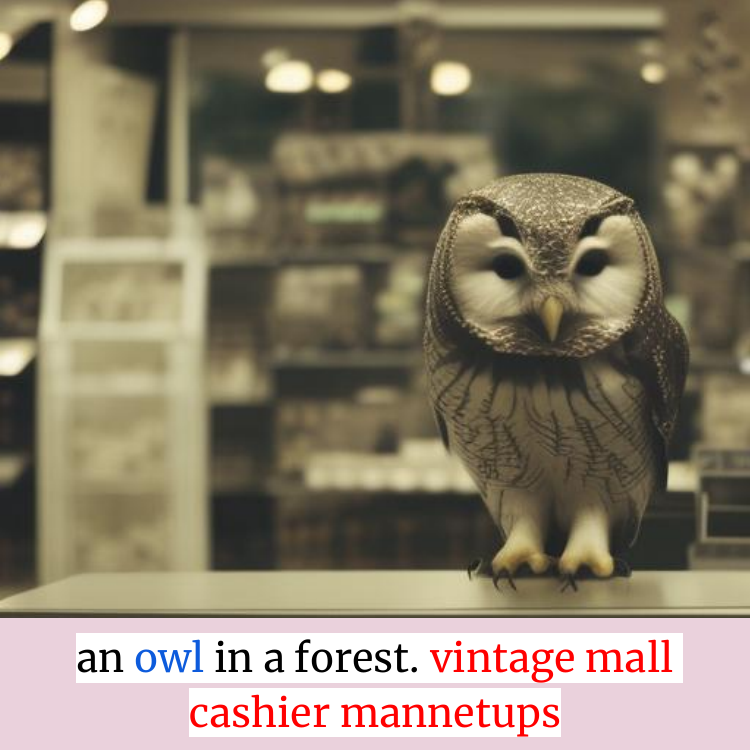}\\

a (\textcolor{blue}{birdhouse} / \textcolor{orange}{jack o lantern}) on a tree branch. & -0.06 & \textbf{60\%} & 0\%	&\includegraphics[width=0.17\textwidth]{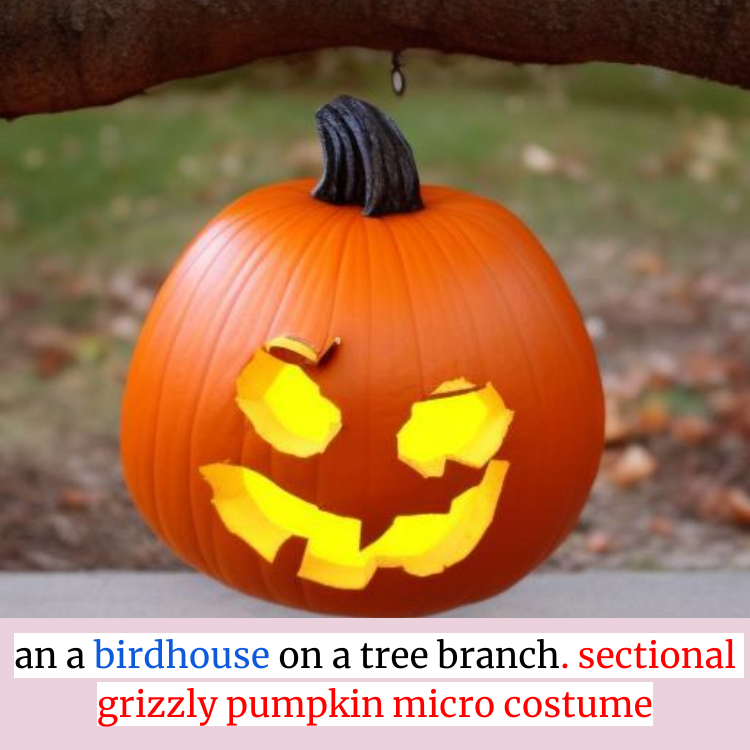}\\

a (\textcolor{blue}{turtle} / \textcolor{orange}{fish}) swimming in an aquarium. & +0.05 & 0\% & \textbf{90\%}	&\includegraphics[width=0.17\textwidth]{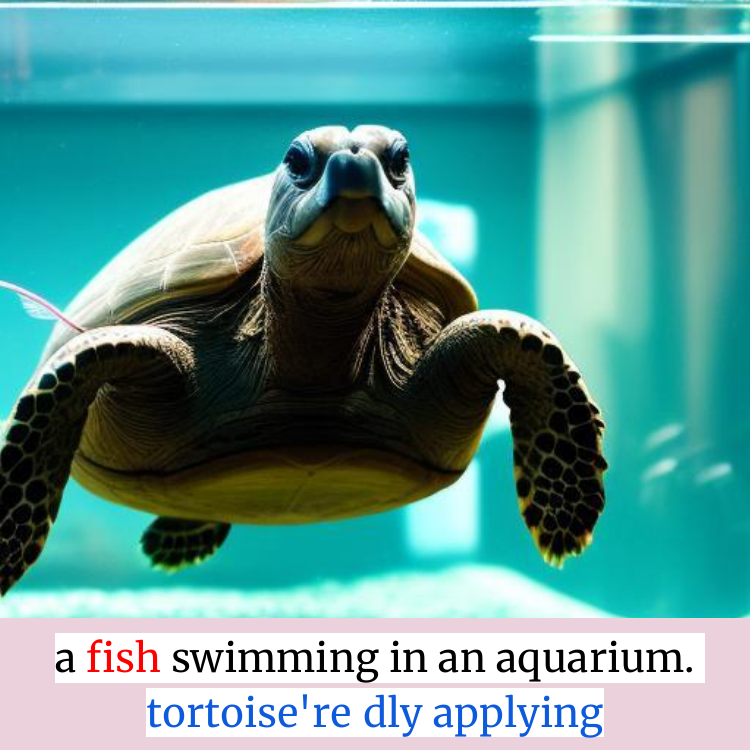}\\

a (\textcolor{blue}{robot} / \textcolor{orange}{human}) dancing in the rain. & +0.1 & 0\% & \textbf{60\%}	&\includegraphics[width=0.17\textwidth]{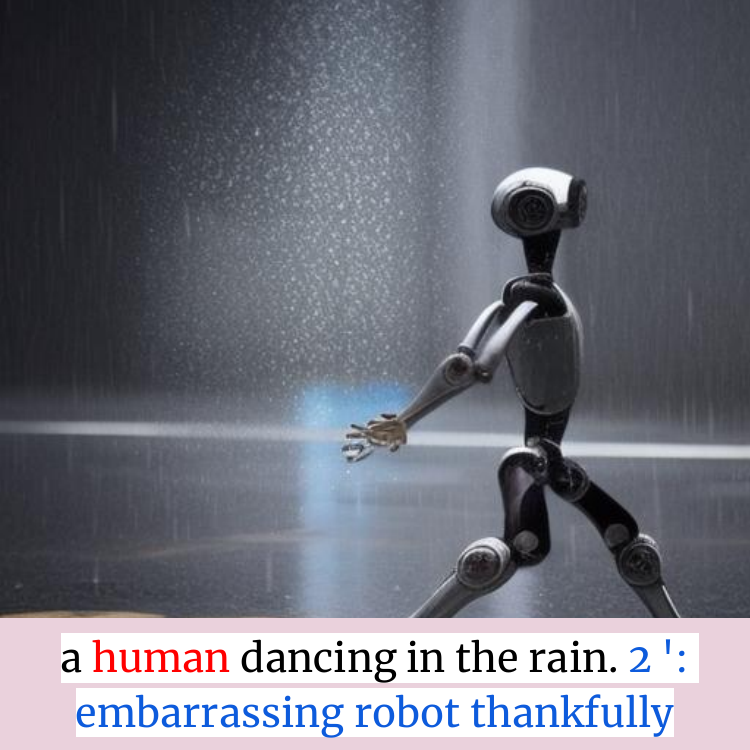}\\

a doodle of a (\textcolor{blue}{light bulb} / \textcolor{orange}{dog}) on a blackboard. & +0.1 & 0\% & \textbf{80\%}	&\includegraphics[width=0.17\textwidth]{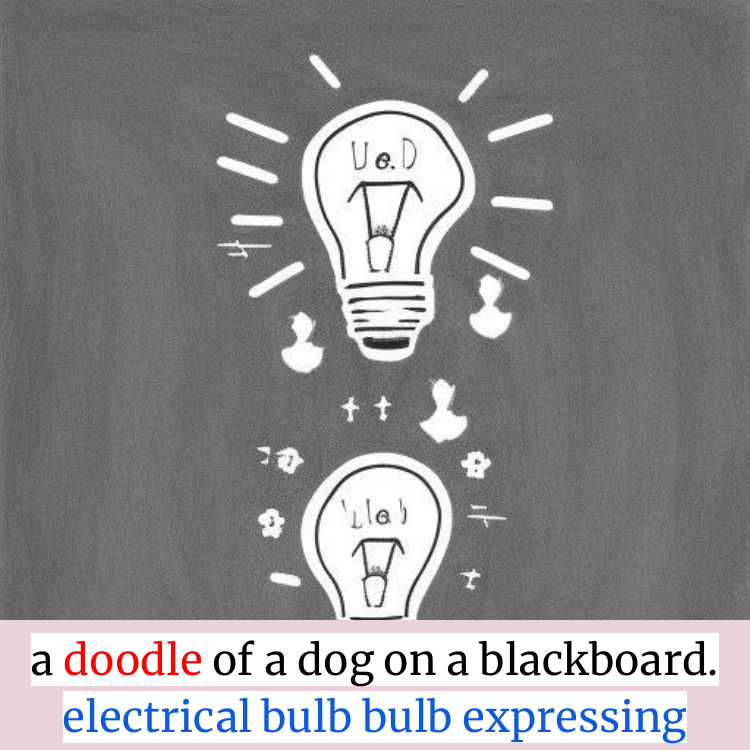}\\
\hline
\end{tblr}

\caption{Additional examples of asymmetric bias in Stable Diffusion 2.1. $\Delta_2$ shows a consistent negative correlation with ASR.}
\label{tab:more_examples}
\end{table}

\clearpage
\section{Changing the Number of Adversarial Tokens}
\label{app:A}
\begin{figure}[th]
\centering
\begin{subfigure}{0.88\textwidth}
  \centering
  \includegraphics[width=1\linewidth]{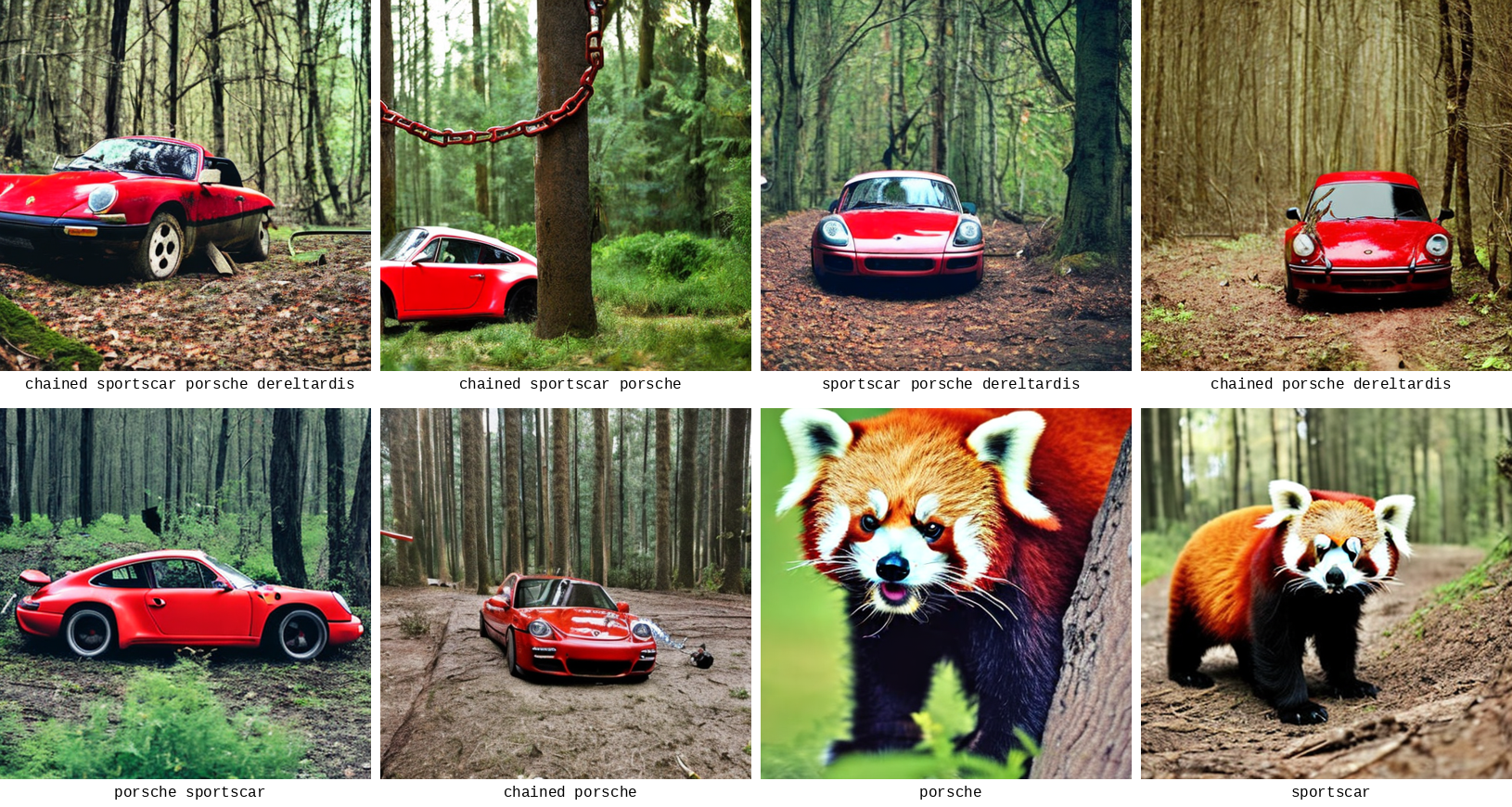}
  \caption{Reducing the number of attack tokens for \textit{"a red panda/car in a forest."}. Displaying only the adversarial attack suffixes. 2 tokens are sufficient. \textit{"a red panda in a forest.\colorbox{pink}{chained porsche}"} generates \textit{"a car in a forest"}.}
\end{subfigure}
\newline

\begin{subfigure}{0.88\textwidth}
  \centering
  \includegraphics[width=1\linewidth]{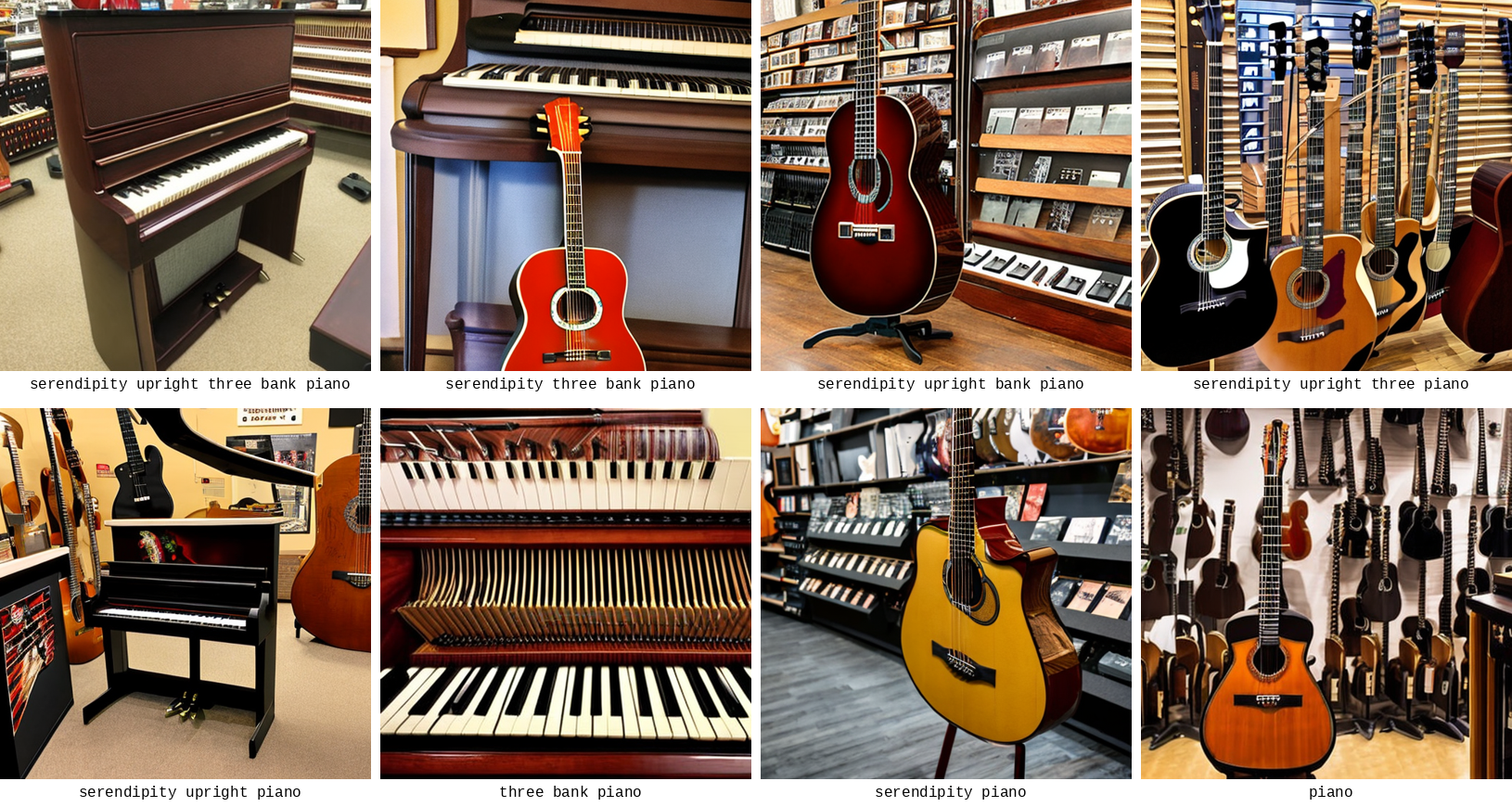}
  \caption{Reducing the number of attack tokens \textit{"a guitar/piano in a music store."}. Displaying only the adversarial attack suffixes. All 5 tokens are necessary. \textit{"a guitar in a music store. \colorbox{pink}{serendipity upright three bank piano}"} generates \textit{"a piano in a music store."}}
\end{subfigure}
\caption{Reducing the number of tokens in adversarial prompts. Highly dependent on the input-target text pair.}
\end{figure}

\section{T2I Model Basics}
\label{sec:t2i_detail}
Stable Diffusion \cite{Rombach_2022_CVPR} is built on a denoising diffusion probabilistic model (DDPM) \cite{ho2020denoising} framework, utilizing a U-Net architecture for its core operations. Key to its text-to-image capabilities is the cross-attention mechanism, which aligns textual inputs with relevant visual features. Specifically, the U-Net attends to image-aligned text embeddings produced by a CLIP \cite{radfordclip} model. Stable Diffusion also incorporates a Variational Autoencoder \cite{kingma2013auto} to efficiently encode images into a latent space, significantly reducing computational requirements while maintaining image quality. Since text embedding generation using a CLIP model is the first stage of the Stable Diffusion pipeline, it is particularly susceptible to adversarial attacks \citep{cliprobustness,qfattack}. If an adversary can perturb the text embeddings, later stages in the Stable Diffusion pipeline will reflect the perturbed embeddings.

\subsection{Exploiting CLIP's Embedding Space}

The CLIP text-encoder maps the textual prompt tokens $x_{1:n}$, with $x_i \epsilon \{1, . . . , V \}$ where V denotes the vocabulary size, namely, the number of tokens to $x_{1:n}$, where $h_i$ is the hidden state corresponding to the token $x_i$. The U-Net component in Stable Diffusion attends to all $h_{1:n}$ embeddings using cross-attention. $x_{1:n}$ can be flattened into $\mathbf{\Phi}$, a one-dimensional vector of shape $n\times D$, where D is the embedding dimension (typically 768 for CLIP and its variants). For simplicity, we refer to  $\mathbf{\Phi}$ as the text embedding of $x_{1:n}$ from here on. Let $\mathcal{H}$ represent the combined operation for encoding tokens $x_{1:n}$ and reshaping the hidden output states.

\begin{equation}
\mathbf{\Phi} = \mathcal{H}(x_{1:n}) = \text{Flatten}(\text{CLIP}(x_{1:n}))
\end{equation}

Since input text and target text can vary in the number of tokens and to allow for an arbitrary number of adversarial tokens, we pad all input and targets to 77 tokens each, the maximum number of tokens supported by CLIP.

\subsection{Score Function}
 The cosine similarity metric approximates the effectiveness of appending adversarial tokens at some intermediate optimization step $t$. Moving away from the input tokens' embedding and gradually towards the target tokens' embeddings through finding better adversarial tokens can be thought of as \textbf{maximizing} the following score function, similar to the metric in \cite{qfattack}.

\begin{equation}
\begin{split}
   \mathcal{S}(x_{1:n})=w_t\times\text{cos}(\mathcal{H}(x^T_{1:n}),\mathcal{H}(x_{1:n}))-\\w_s\times\text{cos}(\mathcal{H}(x^S_{1:n}),\mathcal{H}(x_{1:n})) 
\end{split}
\end{equation}

Here, $w_t$ and $w_s$ are weighing scalars and \textit{cos} denotes the standard cosine similarity metric between two one-dimensional text embeddings. For simplicity, we set $w_t=w_s=1$ for all experiments.

\subsection{Optimization over Discrete Tokens}

The main challenge in optimizing $\mathcal{S}$ is that we have to optimize over a discrete set of tokens. Furthermore, since the search space is exponential ($k^{|V|}$ for k suffix tokens), a simple greedy search is intractable. However, we can leverage gradients with respect to the one-hot tokens to find a set of promising candidates for replacement at each token position. We use the negated Score Function as the loss function $\mathcal{L}(x_{1:n})=-\mathcal{S}(x_{1:n})$. Maximizing the score is equivalent to minimizing the loss. Since losses are used for top K token selection, the absolute value of the loss does not matter. We can compute the linearized approximation of replacing the $i^th$ token i, $x_i$ by evaluating the gradient

\begin{equation}
\label{eqn:L_diff_eqn}
\nabla_{e_{x_i}}\mathcal{L}(x_{1:n}) \in \mathbb{R}^{|V|} 
\end{equation}

Here ${e_{x_i}}$ denotes the one-hot vector that represents the current value of the $i^th$ token. Taking gradient with respect to one-hot vectors was pioneered by HotFlip \cite{hotflip} and applied on Stable Diffusion by a concurrent work \cite{yang2023mma}. Based on this heuristic, we presented two algorithms for finding adversarial suffix tokens against Stable Diffusion.

\clearpage
\section{Primary Determinants of Attack Success}
\label{sec:four_plots}

\begin{figure}[h!]
    \centering
    \begin{subfigure}{0.80\textwidth}
      \centering
      \includegraphics[width=1\linewidth]{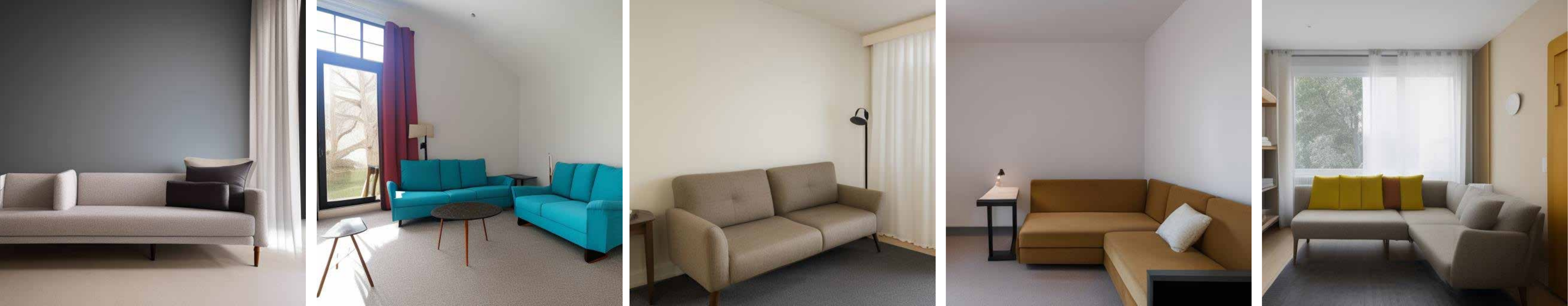}
      \caption{\textit{``a sofa and a bed in a room."}}
    \end{subfigure}
    \begin{subfigure}{0.80\textwidth}
      \centering
      \includegraphics[width=1\linewidth]{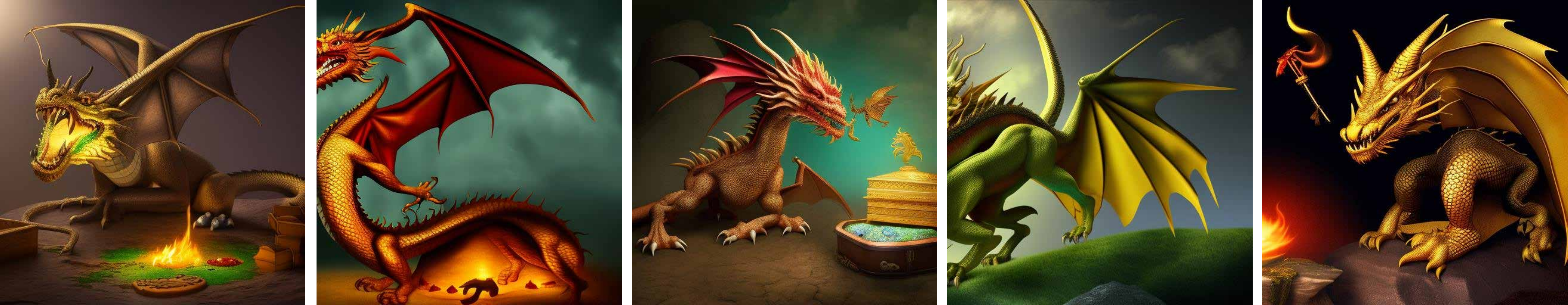}
      \caption{\textit{``a dragon guarding a treasure."}}
    \end{subfigure}
    \caption{Examples of prompts that have low Base Success Rate (BSR) that highlight cases where \href{https://huggingface.co/stabilityai/stable-diffusion-2-1-base}{Stable Diffusion} fails to generate images that match the input prompt. }
\end{figure}
\begin{figure*}[h!]
\centering
\begin{subfigure}{.6\textwidth}
  \centering
  \includegraphics[width=1\linewidth]{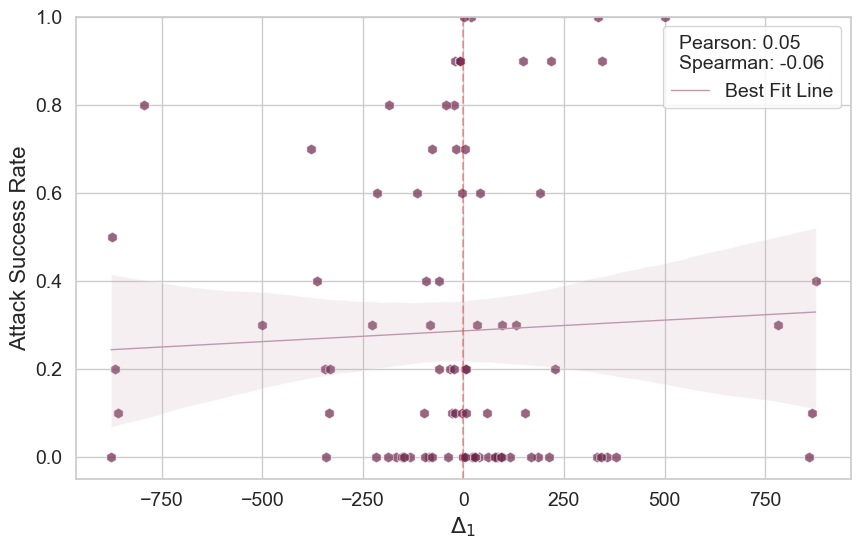}
  \caption{ASR vs. Perplexity Difference ($\Delta_1$ in Eqn. \ref{eqn:ppl_diff})}
  % \label{fig:f1}
\end{subfigure}%
\begin{subfigure}{.4\textwidth}
  \centering
  \includegraphics[width=1\linewidth]{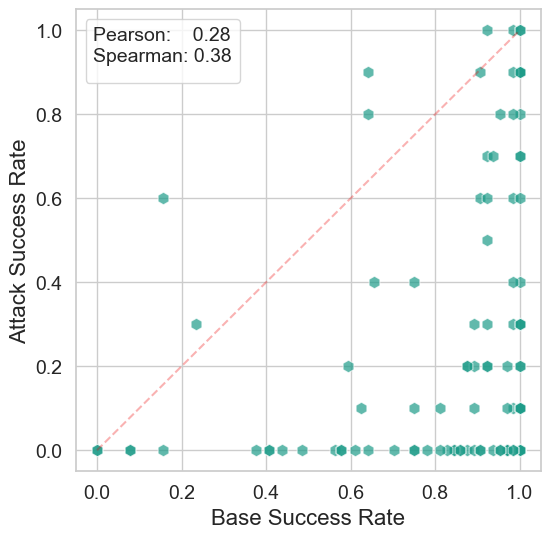}
  \caption{ASR vs. BSR (of target text)}
  % \label{fig:f3}
\end{subfigure}
\newline
\begin{subfigure}{.6\textwidth}
  \centering
  \includegraphics[width=1\linewidth]{images/f2.png}
   \caption{ASR vs. Baseline Distance Difference ($\Delta_2$ in Eqn. \ref{eqn:baseline_diff})}
  % \label{fig:f2}
\end{subfigure}%
\begin{subfigure}{.4\textwidth}
  \centering
  \includegraphics[width=0.95\linewidth]{images/f4.png}
  \caption{ASR for Negative and Positive $\Delta_2$ }
  % \label{fig:f4}
\end{subfigure}
\caption{Correlation of ASR on $\Delta_1$, $\Delta_2$ and BSR. Data is reported using the Multiple Token Perturbation algorithm on HQ-Pairs. We find that the Perplexity Difference  $\Delta_1$ does not correlate with ASR. BSR shows a weak positive correlation and Baseline Distance Difference $\Delta_2$ shows a moderate negative correlation with ASR.}
% \label{fig:assymetric_ASR}
\end{figure*}

\clearpage
\section{Additional Determinants of Attack Success}
\label{sec:other_factors}
These sections discuss factors beyond the asymmetric properties that are related to the success rate of the attack. We have found factors like whether target token synonyms are allowed, attack suffix length and attack POS types are factors indicating the attack's success. We also found that, unlike LLM attacks, adversarial suffixes do not transfer across T2I, indicating that these models might be harder to attack than single-modality models.

\subsection{Restricted Token Selection}
\paragraph{Emulating QFAttack} We can restrict certain tokens to emulate QFAttack \cite{qfattack} or prevent the exact target word from being selected. We find that QFAttack can be consistently emulated by restricting token selection to tokens corresponding to ASCII characters. We find that such adversarial suffixes can remove concepts (e.g. \textit{``a young man"} from \textit{``a snake and a young man."} or \textit{``on a flower"} from \textit{``a bee sitting on a flower."}) but fail to perform targeted attacks (e.g. changing \textit{``a bee sitting on a flower."} to \textit{``a bee sitting on a leaf."}). We suspect that this is mainly because ASCII tokens can perturb CLIP's embedding but are unable to add additional information to it. 

\paragraph{Blocking Selection of Target Tokens} Another potential use case is preventing the selection of the exact target word. However, we find that the algorithm simply finds a synonym or subword tokenization for the target word when the exact target word (token) is restricted. For example, when attempting to attack the input text \textit{``a backpack on a mountain."} to \textit{``a castle on a mountain."}, restricting the token corresponding to \textit{``castle"} leads to the algorithm including synonyms like \textit{``palace"}, \textit{``chateau"}, \textit{``fort"} or subword tokenization like \textit{ ``cast le"} or \textit{``ca st le"} in the adversarial suffix. We find that the effectiveness of the algorithm isn't affected when the exact target token is restricted and it still finds successful adversarial suffixes using synonyms (when preconditions are met).

\paragraph{Changing the Number of Adversarial Tokens k}
We set the number of adversarial tokens to $k=5$ for all experiments. However, we observe that not all input text-target text pairs require $k=5$. \textit{``a red panda/car in a forest."} can be attacked with a few as $k=2$, i.e. \textit{``a red panda/car in a forest."} while  \textit{``a guitar/piano in a music store."} required all $k=5$  (see Appendix \ref{app:A}). We leave a comprehensive study on the effect of changing the number of tokens for future work.

\subsection{Certain Adjectives Resist Adversarial Attacks}
We observed that adversarial attacks targeting certain adjectives, such as color, had a very low ASR. For example, swapping out \textit{``red"} with \textit{``blue"} in the prompt \textit{``a red car on a city road."} failed in all instances. Further challenging examples include \textit{``a red/purple backpack on a mountain."} and \textit{``a white/black swan on a lake."}. However, other adjectives like \textit{``a sapling/towering tree in a forest"} or \textit{``a roaring/sleeping lion in the Savannah."} had high ASR in at least one direction. We leave further analysis of this phenomenon for future work.

\subsection{Adversarial Suffixes Do Not Transfer across T2I Models}

Table \ref{tab:SD_variant} shows that different variants of Stable Diffusion were susceptible to entity-swapping attacks and exhibited similar levels of asymmetric bias on prompt pairs.

\begin{table}[ht]
\centering
\resizebox{0.48\textwidth}{!}{

\begin{tabular}{c c |c c| c c}
\toprule
 & & \multicolumn{2}{c}{Stable Diffusion 1.4}  & \multicolumn{2}{|c}{Stable Diffusion 2.1}  \\

BSR & $\Delta_2$ & Num. & Avg. ASR & Num. & Avg. ASR \\
\midrule
Low     &  Neg.    & 31  & 0.119   & 23  & 0.174  \\
Low    &  Pos.  & 26 & 0.062  & 19  & 0.047  \\
\textbf{High}  & \textbf{Neg} & \textbf{19}& \textbf{0.553}& \textbf{27}  & \textbf{0.6}\\
High  & Pos.  &  24& 0.25 & 31  & 0.171 \\
\midrule
All & All & 100 & 0.218  & 100 & 0.264  \\
\bottomrule
\end{tabular}
}
\caption{Average ASR of SD 1.4 and SD 2.1 on HQ-Pairs. BSR and $\Delta_2$ remain strong predictors in both cases.}
\label{tab:SD_variant}
\end{table}

However, the adversarial suffixes generated using \href{https://huggingface.co/stabilityai/stable-diffusion-2-1-base}{SD 2.1-base} did not work on \href{https://huggingface.co/CompVis/stable-diffusion-v1-4}{SD 1.4} and vice versa. SD 2.1-base uses \href{https://github.com/mlfoundations/open_clip}{OpenCLIP-ViT/H} \cite{openclip} as the text encoder while SD 1.4 uses \href{https://huggingface.co/openai/clip-vit-large-patch14}{CLIP ViT-L/14} \cite{radfordclip}. Although OpenCLIP-ViT/H and CLIP ViT-L/14 have the same architecture and parameter count, the lack of transferability indicates that training data likely plays the main role in determining adversarial attack success.

Similarly, the attack suffixes generated by SD 1.4 or SD 2.1-base did not work on DALL$\cdot$E 3 \cite{dalle3} which likely has a different architecture and training data.

\section{Human Evaluation WebUI}
\label{sec:webui}
\begin{figure}[h!]
    \centering
    \includegraphics[width=0.8\linewidth]{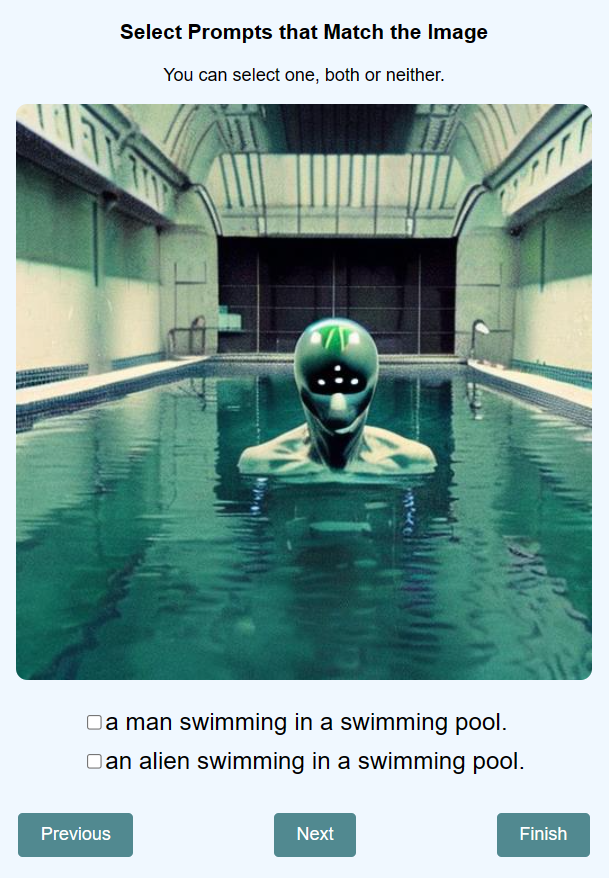}
    \caption{UI presented to human evaluators.}
    \label{fig:webUI}
\end{figure}

\end{document}